\pdfoutput=1
\documentclass[11pt]{article}
\PassOptionsToPackage{numbers,sort&compress}{natbib}
\usepackage[preprint]{acl}
\usepackage{times}
\usepackage{latexsym}
\usepackage[T1]{fontenc}
\usepackage[utf8]{inputenc}
\usepackage{microtype}
\usepackage{graphicx}
\usepackage{booktabs}
\usepackage{array}
\usepackage{amsmath}
\usepackage{xcolor}
\setcitestyle{numbers,square,comma}
\newcommand{\jlauroc}{.974}

\newcommand{\jlece}{.052}

\newcommand{\jlbrier}{.061}

\newcommand{\ntest}{577}
\newcommand{\ntestscen}{41}

\newcommand{\dmthreeauroc}{+.027}
\newcommand{\dmthreeaurocci}{[$-$.011, .065]}

\newcommand{\dmthreeece}{$-$.002}
\newcommand{\dmthreeececi}{[$-$.048, .036]}

\newcommand{\dmbauroc}{+.036}
\newcommand{\dmbaurocci}{[$-$.009, .114]}

\newcommand{\mbece}{.027}

\newcommand{\dzsauroc}{+.283}
\newcommand{\dzsaurocci}{[.186, .392]}

\newcommand{\zsece}{.170}

\newcommand{\dmthreelegit}{+.177}
\newcommand{\dmthreelegitci}{[.008, .360]}
\newcommand{\mthreefamLegit}{.823}

\newcommand{\mthreediscauroc}{.823}

\newcommand{\nimargin}{.009}
\newcommand{\latpfifty}{64.5}
\newcommand{\latpninetyfive}{110.5}
\newcommand{\latextra}{21.8}

\newcommand{\latfivefresh}{321.6}
\newcommand{\latfiveshared}{168.7}
\newcommand{\latdrift}{6/1500}
\newcommand{\speedup}{30}
\newcommand{\speeduppninetyfive}{52}
\newcommand{\mblatpfifty}{29.7}
\newcommand{\poljltau}{0.92}
\newcommand{\polmtau}{0.97}
\newcommand{\poljlscam}{40/52}

\newcommand{\polmscam}{41/52}

\newcommand{\dturn}{1.14}
\newcommand{\dturnci}{[0.55, 1.67]}
\newcommand{\dturnboth}{36}
\newcommand{\dturnearlier}{25}
\newcommand{\dturnsame}{6}
\newcommand{\dturnlater}{5}
\newcommand{\stTurnEnd}{.927}
\newcommand{\stMid}{.888}

\newcommand{\stGap}{3.8}
\newcommand{\stGapci}{[2.2, 5.6]}
\newcommand{\turnone}{.854}
\newcommand{\turntwo}{.927}
\newcommand{\swapflip}{2.6}
\newcommand{\swapflipci}{[0.2, 5.8]}
\newcommand{\paraflip}{5.2}
\newcommand{\paraflipci}{[2.0, 9.4]}
\newcommand{\paratolegit}{28}
\newcommand{\paran}{30}
\newcommand{\uqAmount}{+.013}
\newcommand{\uqAmountci}{[.002, .033]}
\newcommand{\uqAmountjl}{.996}
\newcommand{\uqAmountzs}{.984}
\newcommand{\uqPhone}{+.041}
\newcommand{\uqPhoneci}{[.020, .075]}
\newcommand{\uqPhonejl}{.999}
\newcommand{\uqPhonezs}{.957}
\newcommand{\uqUrgency}{$-$.062}
\newcommand{\uqUrgencyci}{[$-$.133, .013]}
\newcommand{\uqUrgencyjl}{.863}
\newcommand{\uqUrgencyzs}{.925}
\newcommand{\uqSympathy}{$-$.113}
\newcommand{\uqSympathyci}{[$-$.230, .001]}
\newcommand{\uqSympathyjl}{.447}
\newcommand{\uqSympathyzs}{.560}

\newcommand{\seedaurocmin}{.969}
\newcommand{\seedaurocmax}{.977}
\newcommand{\seedcorr}{.987}

\newcommand{\nboot}{2000}
\newcommand{\armAlabauroc}{.875}
\newcommand{\armAlabacc}{.892}
\newcommand{\armAlabece}{.107}

\newcommand{\armAlablegit}{1.000}
\newcommand{\armAlabgraylegit}{.095}
\newcommand{\armAlabms}{317}
\newcommand{\armAverbauroc}{.914}
\newcommand{\armAverbacc}{.896}
\newcommand{\armAverbece}{.069}

\newcommand{\armAverblegit}{1.000}
\newcommand{\armAverbgraylegit}{.095}

\newcommand{\armAfirstauroc}{.950}
\newcommand{\armAfirstacc}{.892}
\newcommand{\armAfirstece}{.050}

\newcommand{\armAfirstlegit}{1.000}
\newcommand{\armAfirstgraylegit}{.095}

\newcommand{\armBauroc}{.972}
\newcommand{\armBacc}{.894}
\newcommand{\armBece}{.050}

\newcommand{\armBlegit}{1.000}
\newcommand{\armBgraylegit}{.095}

\newcommand{\armBceauroc}{.970}
\newcommand{\armBceacc}{.910}
\newcommand{\armBceece}{.076}

\newcommand{\armBcelegit}{.994}
\newcommand{\armBcegraylegit}{.095}

\newcommand{\armCauroc}{.962}
\newcommand{\armCacc}{.927}
\newcommand{\armCece}{.055}

\newcommand{\armClegit}{1.000}
\newcommand{\armCgraylegit}{.429}
\newcommand{\armCms}{80}
\newcommand{\armAparsefail}{0}
\newcommand{\armAlabseedauroc}{.875 / .878}
\newcommand{\armAlabseedacc}{.892 / .894}
\newcommand{\armAlabseedece}{.107 / .106}

\newcommand{\armAfirstseedauroc}{.950 / .955}
\newcommand{\armAfirstseedacc}{.892 / .892}
\newcommand{\armAfirstseedece}{.050 / .036}

\newcommand{\armBseedauroc}{.972 / .967}
\newcommand{\armBseedacc}{.894 / .931}
\newcommand{\armBseedece}{.050 / .021}
\newcommand{\armBseedgraylegit}{.095 / .405}
\newcommand{\armBceseedauroc}{.970 / .966}
\newcommand{\armBceseedacc}{.910 / .905}
\newcommand{\armBceseedece}{.076 / .051}

\newcommand{\armCseedauroc}{.962 / .971}
\newcommand{\armCseedacc}{.927 / .922}
\newcommand{\armCseedece}{.055 / .037}
\newcommand{\armCseedgraylegit}{.429 / .548}

\newcommand{\armAnoconf}{0}
\newcommand{\armTgenlab}{4.06}
\newcommand{\armTbce}{5.24}
\newcommand{\armTb}{3.88}
\newcommand{\armThead}{5.18}
\newcommand{\armBceeceCapped}{.065}
\newcommand{\armCeceCapped}{.042}
\newcommand{\armAlatratio}{4.9}

\newcommand{\mbauxece}{.042}
\newcommand{\mbauxbrier}{.069}

\newcommand{\mbauxdelta}{+.026}
\newcommand{\mbauxdeltaci}{[$-$.023, .105]}
\newcommand{\mbauxT}{2.9}
\newcommand{\mbauxepoch}{2}

\newcommand{\dgenauroc}{+.024}
\newcommand{\dgenaurocci}{[$-$.008, .074]}
\newcommand{\dgenlabauroc}{+.098}
\newcommand{\dgenlabaurocci}{[.021, .196]}
\newcommand{\mthreeTfit}{1.49}
\newcommand{\mthreeTece}{.049}
\newcommand{\mthreeTbrier}{.096}

\newcommand{\eceDiffMthreeT}{+.003}
\newcommand{\eceDiffMthreeTci}{[$-$.056, .030]}
\newcommand{\seedAdelta}{+.031}
\newcommand{\seedAdeltaci}{[$-$.005, .067]}

\newcommand{\seedBdelta}{+.023}
\newcommand{\seedBdeltaci}{[$-$.018, .063]}

\newcommand{\seedCdelta}{+.024}
\newcommand{\seedCdeltaci}{[$-$.018, .064]}

\newcommand{\seedsPassing}{all three}

\newcommand{\recipeAseedmean}{.930}
\newcommand{\recipeDseedmean}{.972}

\newcommand{\ruleEligible}{two}
\newcommand{\ruleEligibleNames}{the intent-question recipe (i7) and the two-epoch LoRA recipe (lora\_r16)}

\newcommand{\valPickAcc}{.888}
\newcommand{\valPickAltAcc}{.865}

\newcommand{\armBmsMeasured}{63.6}

\newcommand{\armBmsMeasuredExtra}{20.8}

\newcommand{\turnDiffCiUnit}{scenarios}

\newcommand{\vertorch}{2.10.0+cu128}
\newcommand{\vertransformers}{4.57.6}
\newcommand{\verpeft}{0.19.1}
\newcommand{\verpython}{3.12.3}
\newcommand{\testgenseed}{not recorded}
\newcommand{\jlaurocciPct}{[93.3, 99.7]}

\newcommand{\recipeAseedmeanPct}{93.0}
\newcommand{\recipeDseedmeanPct}{97.2}
\newcommand{\recipegainPp}{+3.8}

\title{Open-Jev Judgments on CallScreenBench:\\
Calibrated One-Pass Scam Screening with a Small Language Model}

\author{Simiao Ren$^{\dagger}$ \quad Kidus Zewde$^{*}$ \quad Xingyu Shen$^{*}$ \quad Yuchen Zhou$^{*}$ \quad Dennis Ng$^{*}$ \\
\textbf{Ankit Raj}$^{*}$ \quad \textbf{Tommy Duong}$^{*}$ \quad \textbf{Yuxin Zhang}$^{*}$ \quad \textbf{Neo Tiangratanakul}$^{*}$ \\
Scam.ai (Reality Inc.) \\
$^{*}$Equal contribution. \quad $^{\dagger}$Corresponding author: \texttt{benren@scam.ai}}

\begin{document}
\maketitle
{\renewcommand{\thefootnote}{}\footnote{Authorship of this paper does not imply employment by or affiliation with Scam.AI (Reality Inc.); some authors are independent researchers who collaborated on this work.}\addtocounter{footnote}{-1}}

\begin{abstract}
Screening a phone call for fraud needs a trustworthy probability after every caller turn, in milliseconds. Jev-style typed decisions promise exactly that: declared options go in, one calibrated probability per option comes out of a single forward pass, with no generated text. We test an open implementation of this readout, JevLite, on scam-call screening: Qwen3-4B is LoRA-tuned so that the temperature-scaled softmax over two answer-label logits is $P(\text{scam})$. On \ntestscen{} held-out CallScreenBench scenarios (\ntest{} per-turn decisions) a three-seed ensemble reaches AUROC \jlauroc{} with calibration error \jlece{}, non-inferior to an LLM judge (MiniMax-M3) at a pre-registered .02 margin, with no false alarms on legitimate calls, decisions \dturn{} turns earlier under the same hang-up rule, and \latpfifty{} ms per decision on one consumer GPU, \armAlatratio$\times$ lower than the same backbone fine-tuned to generate its answer. The gain is in the readout and calibration, not accuracy: a fine-tuned ModernBERT encoder is not significantly worse, the recipe was selected with test-set exposure, and all callers are synthetic. We claim no architectural novelty; the contribution is the application and an evaluation reporting calibration, false alarms and decision timing alongside AUROC.
\end{abstract}

\section{Introduction}

Phone scams are increasingly screened by software that answers on the user's behalf, from on-device assistants to research prototypes of delegated ``call secretaries'' \citep{pandit2023robocall,callscreenbench2026}. The callers are increasingly automated too: voice-enabled LLM agents carry out common scams end to end \citep{fang2024voiceagents}, more than a quarter of unwanted calls in a 66-day honeypot deployment opened with a recorded or synthetic voice \citep{own_machines2026}, and real scam calls reveal their intent early \citep{anatomy2026}. A screener must therefore reach a judgment that software can act on---hang up, take a message, or put the call through---early in the call, within hundreds of milliseconds, and without hanging up on the user's bank or doctor \citep{callscreenbench2026}.

\begin{figure*}[t]
  \centering
  \includegraphics[width=\textwidth]{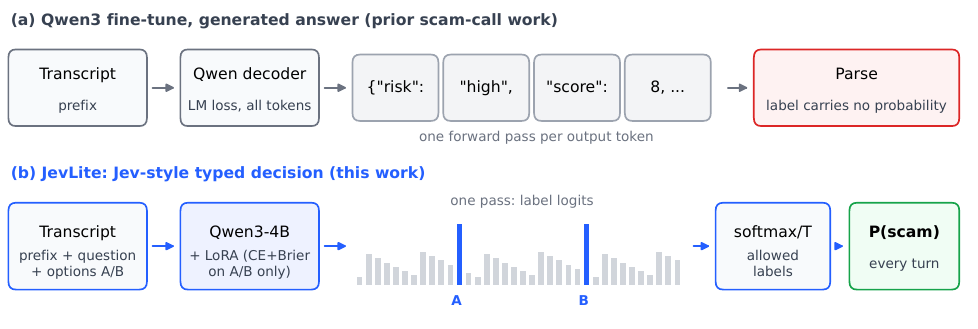}
  \caption{(a) Prior scam-call classifiers fine-tune a decoder with a language-modeling loss over a generated answer and parse the output. (b) JevLite uses the same kind of decoder but restricts training and readout to the declared answer labels: one forward pass yields a softmax over the labels, which a temperature fitted on validation data turns into a calibrated $P(\text{scam})$ after every caller turn.}
  \label{fig:overview}
\end{figure*}

Current scam-call work either prompts a large LLM to classify the call \citep{shen2024combating,shen2025warned} or fine-tunes a small decoder to \emph{generate} a structured answer that is then parsed \citep{ghafariasl2026elder,wang2026safeqaq}. Both treat the decision as text: generated scores are not calibrated probabilities unless trained to be \citep{lin2022words,stangel2025rewardingdoubt,conftuner2025}, small models can fail to produce parseable output \citep{ghafariasl2026elder}, and calibration and false alarms on legitimate calls are rarely reported (Section~\ref{sec:related}).

Jev \citep{typesafe2026jev} replaces generated text with typed decisions---a state and questions with declared options go in, one probability per option comes out---and open reimplementations obtain the same interface from ordinary decoders by reading answer-label logits in one forward pass \citep{semif2026,openjevsglang2026}. Is that interface a good fit for call screening, where decisions must be fast, repeated at every turn, and trustworthy as probabilities?

We make no architectural claim: JevLite follows the open reimplementations, and label-token readout itself is established \citep{schick2021pet,zhao2021calibrate,inan2023llamaguard}. Our contribution is the application and its evaluation (Figure~\ref{fig:overview}): we (i) adapt the readout to per-turn scam screening, fine-tuning it with teacher-labeled intent questions; (ii) evaluate it on held-out CallScreenBench scenarios against an LLM judge \citep{zheng2023judge}, a fine-tuned ModernBERT-large encoder \citep{warner2024modernbert}, a Qwen3 fine-tune of the same backbone and the untuned backbone, reporting calibration, false alarms on legitimate calls and decision timing alongside AUROC, with scenario-level confidence intervals and seed replication; and (iii) report where the approach fails: ambiguous calls, mid-utterance decisions, prompt sensitivity, and a recipe selection that was not blind to the test set.

\section{Related Work}
\label{sec:related}

\paragraph{Scam-call detection.} Datasets range from synthetic English dialogues, which keyword baselines nearly solve \citep{shen2024combating}, to Chinese audio--text corpora \citep{teleantifraud2025,wang2026safeqaq}, streaming labels for Indian-English calls \citep{icfd2026}, and real honeypot calls \citep{anatomy2026}. Methods range from prompted judges with per-utterance warnings \citep{shen2025warned} and reinforcement-learned audio LLMs \citep{wang2026safeqaq} to turn-by-turn tactic prediction \citep{vishingtactics2026} and instruction-tuned small models that re-estimate risk each turn \citep{ghafariasl2026elder}. These works report accuracy or F1; none of the LLM-based ones reports calibration, and false alarms on legitimate callers are rarely measured. LLM-paraphrased attacks on vishing classifiers \citep{li2025phisher} and automated voice-agent scams \citep{fang2024voiceagents} show that the threat itself is adapting.

\paragraph{LLMs as classifiers and calibration.} Scoring answer labels with next-token probabilities predates instruction tuning \citep{brown2020gpt3,schick2021pet,robinson2023mcq}; restricting the softmax to the allowed labels and correcting the residual label bias is contextual calibration \citep{zhao2021calibrate,holtzman2021surface}, and Llama Guard and ShieldGemma publish an answer token's probability as their score \citep{inan2023llamaguard,zeng2024shieldgemma}. Whether a model's stated confidence can be trusted is contested \citep{tian2023justask,xiong2024uncertainty,kadavath2022}; first-token probabilities can disagree with the generated answer \citep{wang2024myanswerisc}, option order changes answers \citep{zheng2024mcq,pezeshkpour2024order}, and instruction tuning pushes answer-token scores towards over-confidence \citep{cruz2024letters}. Calibration is measured with expected calibration error \citep{naeini2015ece,nixon2019measuring} and proper scoring rules \citep{brier1950,gneiting2007proper}, restored post hoc by temperature scaling \citep{guo2017calibration} or trained for directly \citep{lin2022words,stangel2025rewardingdoubt,conftuner2025}. Distilling a large judge into a small model \citep{kim2024prometheus,hsieh2023distilling}, fine-tuned encoders for classification \citep{warner2024modernbert,ettin2025}, early risk detection on text streams \citep{losada2016erisk} and selective prediction \citep{geifman2017selective} are the other threads we draw on.

\paragraph{Jev and open reimplementations.} The vendor describes Jev as a closed ``System One'' model with a ``parallel sampler'', trained by ``reinforcement learning for calibrated decisions'' \citep{typesafe2026jev}; none of this is public, and no independent evaluation of Jev exists at the time of writing. Open attempts rebuild the interface on frozen decoders \citep{semif2026,openjevsglang2026}, diffusion language models \citep{vllm57250}, fine-tuned decoders \citep{nimble2026} and encoders \citep{laya2026}; reading answer letters matches grammar-constrained JSON, while model-written probabilities collapse \citep{minijev2026}. None has been evaluated on scam calls or on calibrated per-turn decisions.

\section{Method}

\paragraph{Typed decisions.} A decision is a question $q$ with $K$ declared options, each rendered with a single-token label $\ell_1,\dots,\ell_K$ (A, B, \ldots). Given a transcript prefix $x$, the prompt $(x, q, \text{options})$ (Appendix~\ref{app:prompts}) is run through the decoder once, and the logits $z_k$ of the label tokens at the last position are normalized over the allowed labels only \citep{zhao2021calibrate,holtzman2021surface}, with a temperature $T$ fitted on validation data (below):
\begin{equation}\label{eq:softmax}
p_k = \frac{\exp(z_k/T)}{\sum_{j=1}^{K}\exp(z_j/T)} .
\end{equation}
No text is generated, so the output is always one of the declared options. For the main question, \emph{is this caller attempting a scam or fraud rather than a legitimate call?}, $K=2$ and $P(\text{scam}) = p_{\text{A}}$.

\paragraph{Three ways to read the same decoder.} Every fine-tuned Qwen3 system in this paper is Qwen3-4B with LoRA adapters; what differs is the training target, the position that is read, and whether decoding happens (Table~\ref{tab:readouts}). (i) \emph{Parsed label}, as in prior scam-call work: next-token loss over a JSON answer, which is generated token by token and parsed; the output can be malformed and carries no probability, since a number written into the text is more generated text. (ii) \emph{Label-token readout}: the same weights, but the answer prefix \texttt{\{"scam": "} is teacher-forced and the renormalized, temperature-scaled probabilities of the \texttt{yes}/\texttt{no} tokens are read in one pass; the training objective never targeted this distribution directly. (iii) \emph{Typed decision} (JevLite, following Jev): the options are declared as single-token labels and the loss (cross-entropy plus Brier) is applied only to those label logits, normalized over the allowed labels (Eq.~\ref{eq:softmax}). The quantity trained is exactly the quantity read: one forward pass, an output restricted to the declared options by construction, and a probability whose calibration is part of the objective.

\begin{table*}[t]
\centering
\small
\setlength{\tabcolsep}{4pt}
\renewcommand{\arraystretch}{1.15}
\begin{tabular}{@{}>{\raggedright\arraybackslash}p{2.4cm}>{\raggedright\arraybackslash}p{2.7cm}>{\raggedright\arraybackslash}p{2.9cm}>{\raggedright\arraybackslash}p{1.7cm}>{\raggedright\arraybackslash}p{2.3cm}>{\raggedright\arraybackslash}p{2.6cm}@{}}
\toprule
\textbf{Readout} & \textbf{Training loss} & \textbf{At inference} & \textbf{Forward passes} & \textbf{Output space} & \textbf{Probability} \\
\midrule
Parsed label & next-token loss over the whole answer text & generate the JSON answer, then parse it & one per generated token ($\sim$8) & free text; may be malformed & none (a written number is more text) \\
Label-token readout & same weights as above & teacher-force \texttt{\{"scam": "}, read the next-token distribution & one & \texttt{yes}/\texttt{no} tokens by renormalization & post hoc: renormalize, then fit $T$ on validation \\
Typed decision (JevLite) & CE $+$ Brier on the label logits only, normalized over the declared options & read the label logits at the question & one & declared options, by construction & trained directly; $T$ fitted on validation \\
\bottomrule
\end{tabular}
\caption{Three ways to read one fine-tuned decoder. All three share the weights (Qwen3-4B with LoRA), the transcript rendering and the question; they differ only in what the loss targets, which position is read, and whether the model must decode. Prior scam-call classifiers use the first row; JevLite and the open Jev reimplementations use the third; the second is how the first row's model can be read without changing its training.}
\label{tab:readouts}
\end{table*}

\paragraph{Training.} We attach LoRA adapters \citep{hu2022lora} (rank 16, $\alpha=32$, all attention and multi-layer perceptron (MLP) projections; Appendix~\ref{app:prompts}) to Qwen3-4B \citep{qwen3} and train on the label logits only, with the sum of cross-entropy (CE) and the Brier score over the allowed-label softmax ($T=1$ during training). Every caller-turn prefix of a training call is a training example labeled with the call's ground truth, so the model is trained to answer with the evidence available so far. We add two auxiliary intent questions---whether the caller pushes their own callback channel or discourages independent verification, and whether they ask for a secret such as a PIN or one-time code---whose labels come from MiniMax-M3 as a teacher on training and validation prefixes only, discarding teacher probabilities between .3 and .7. Gray scenarios whose ground truth is legitimate receive a loss weight of 2. The recipe trains for one epoch at learning rate $5\times10^{-5}$.

\paragraph{Calibration and aggregation.} One temperature $T$ per model is fitted by minimizing negative log-likelihood on validation prefixes. We train three models that differ only in seed and average their temperature-scaled $P(\text{scam})$; this ensemble is what Table~\ref{tab:main} calls JevLite. The seeds' logits correlate at $\geq$\seedcorr{} and their AUROCs span \seedaurocmin{}--\seedaurocmax{}. A deployed screener would use a single model, and all latencies refer to one model.

\paragraph{Streaming use.} Screening re-asks the question after every caller turn. Because questions about the same call share the transcript prefix, the prefix is encoded once and each question adds only its short suffix to the cached state.

\begin{figure}[t]
  \centering
  \includegraphics[width=\columnwidth]{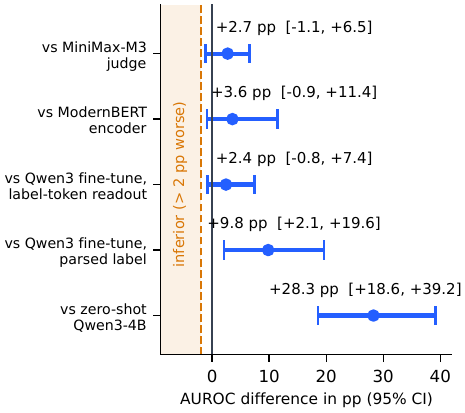}
  \caption{Paired AUROC differences in percentage points (pp), with 95\% scenario-bootstrap intervals. A difference below the dashed line (2 pp worse) would fail the pre-registered non-inferiority test. The Qwen3 fine-tune is the same backbone trained with the ordinary next-token loss to write a JSON answer (arm A), read either by parsing its label or from its label-token probability.}
  \label{fig:forest}
\end{figure}

\begin{figure*}[t]
  \centering
  \includegraphics[width=\textwidth]{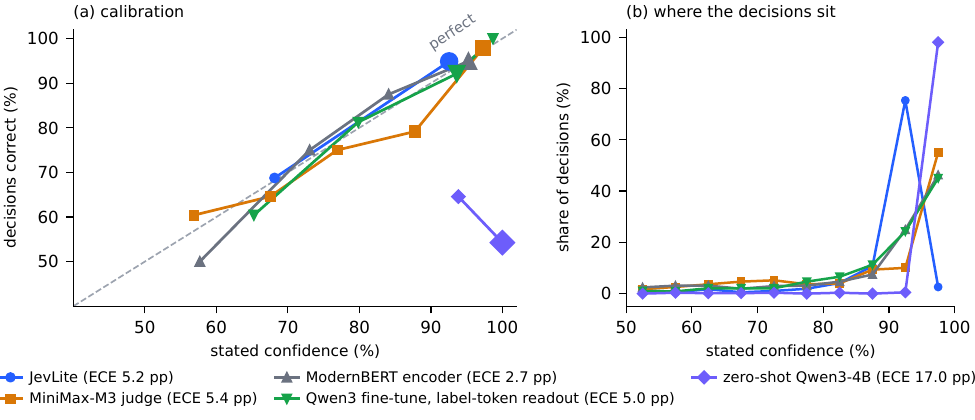}
  \caption{(a) Calibration on the test set, in percentages, including the Qwen3 fine-tune of the same backbone read from its label-token probability (arm A). Bins hold equal numbers of decisions and are merged until no two points lie within 5 percentage points of each other; marker area grows with the decisions behind each point, and points below the diagonal are over-confident. (b) How the decisions are distributed over stated confidence: all trained systems concentrate above 90\%, which is why bin choice matters \citep{nixon2019measuring}. ECE in Table~\ref{tab:main} uses the conventional equal-width bins.}
  \label{fig:reliability}
\end{figure*}

\section{Experimental Setup}
\label{sec:setup}

\paragraph{Data.} CallScreenBench \citep{callscreenbench2026} provides 48 core scenarios (24 scam, 15 legitimate, 9 gray, where gray scenarios have a designated scam or legitimate ground truth) with transcripts of an LLM caller (MiniMax-Text-01) talking to 16 different secretary agents. We split the core scenarios by scenario into 40 for training (3{,}187 turn-prefix decisions) and 8 for validation (650); the validation set is small and underpins every temperature and threshold in the paper. For testing we generate new calls for the 41 evaluation scenarios that do not appear in the core set, with the benchmark's own caller and two scripted secretaries (one that stalls, one that complies), giving 82 calls and \ntest{} turn-end decisions. No test scenario occurs in training or validation. Test scenarios are new instances of the benchmark's scam archetypes, not new archetypes (Appendix~\ref{app:data}).

\paragraph{Baselines.} (1) \emph{The LLM judge}: MiniMax-M3, the judge of our voice-honeypot pipeline \citep{anatomy2026}, prompted on the same prefixes to return a probability in JSON following TypeSafe's official LLM adapter \citep{typesafe_adapter}; its prompt was not tuned, and a discrete yes/no variant reaches only AUROC \mthreediscauroc{}. (2) \emph{The encoder}: ModernBERT-large \citep{warner2024modernbert} fine-tuned as a binary classifier on the same training prefixes, epoch and temperature chosen on validation. (3) \emph{The encoder with auxiliary heads}: the same encoder with two extra heads on the teacher-labeled intent questions JevLite receives, so that supervision is matched. (4) \emph{The Qwen3 fine-tune}: the same backbone trained with the ordinary next-token loss to write \texttt{\{"scam": "yes"\}}, read by parsing or from its label token (Table~\ref{tab:readouts}, Section~\ref{sec:arms}). (5) \emph{Zero-shot}: untuned Qwen3-4B read out exactly as JevLite.

\paragraph{Metrics.} We report AUROC, accuracy at $P=.5$, expected calibration error \citep[ECE; ten equal-width bins,][]{naeini2015ece} and Brier score \citep{brier1950,gneiting2007proper}, plus accuracy by call family. All trained systems and the zero-shot backbone are temperature-scaled on validation; the judge's verbalized probability is used as emitted and, in the row so marked, temperature-scaled as well. Intervals are 95\% percentile intervals from \nboot{} bootstrap resamples of test \emph{scenarios}, since decisions within a scenario are correlated; paired differences are resampled jointly (Appendix~\ref{app:verify}), and the turn difference resamples \turnDiffCiUnit. ECE intervals are upward-biased at this sample size, so only paired ECE differences are reported; no multiplicity correction is applied. The success criterion was written down before any training run: ``beats the judge on AUROC and ECE, or matches AUROC within 0.02 with at least 20$\times$ lower latency''; the .02 margin is a convention, not derived from an operating-point cost. The superiority arm was not met, so we report the non-inferiority arm, a lower confidence bound on $\Delta$AUROC above $-.02$; the latency clause is met against the judge's API deployment, not against a local Qwen3 fine-tune of the same backbone (\armAlatratio$\times$). Latency is the per-decision median (p50) and 95th percentile (p95) on one RTX~3090 at batch size 1, with no other job running for the JevLite measurement; the judge's is remote API wall time under 8 concurrent requests, including about 108 reasoning tokens per decision.

\paragraph{Model selection.} About ten recipes had been scored on the test set before the selection rule was written: ``best mean test AUROC across at least two seeds among recipes with legitimate-call accuracy at least .97''. Only \ruleEligible{} recipes (\ruleEligibleNames) were eligible; the statistic is the mean single-seed test AUROC (\recipeDseedmean{} against \recipeAseedmean), not the ensemble AUROC of Table~\ref{tab:ablation}. The headline recipe is therefore chosen with test-set exposure and its point estimate is subject to winner's-curse inflation. On the 8-scenario validation set, accuracy favored seed 0 of the chosen recipe (\valPickAcc{} against \valPickAltAcc); Table~\ref{tab:ablation} lists validation AUROC and accuracy for every recipe scored.

\section{Results}
\label{sec:results}

\begin{table*}[t]
\centering
\small
\setlength{\tabcolsep}{1.6pt}
\footnotesize
\begin{tabular}{@{}lccccccc@{}}
\toprule
 & \textbf{JevLite} & \textbf{LLM judge} & \textbf{ModernBERT} & \textbf{ModernBERT} & \textbf{Qwen3 FT} & \textbf{Qwen3 FT} & \textbf{Zero-shot} \\
 & (3-seed ens.) & (MiniMax-M3) & & + aux heads & parsed label & label-token & \\
\midrule
AUROC (\%) & \textbf{97.4} & 94.7 & 93.8 & 94.8 & 87.5 & \underline{95.0} & 69.1 \\
Accuracy (\%) & \textbf{92.7} & 87.0 & 88.9 & \underline{90.6} & 89.2 & 89.2 & 55.1 \\
ECE (\%) $\downarrow$ & 5.2 & 5.4 & \textbf{2.7} & \underline{4.2} & 10.7 & 5.0 & 17.0 \\
\ ECE, judge $T$-scaled (\%) & -- & 4.9 & -- & -- & -- & -- & -- \\
Brier ($\times$100) $\downarrow$ & \textbf{6.1} & 9.5 & 8.2 & \underline{6.9} & 10.7 & 8.1 & 28.0 \\
\multicolumn{8}{@{}l}{\ JevLite, mean of 3 single seeds: AUROC 97.2\%, accuracy 92.3\%, ECE 4.7\%, Brier 6.3} \\
\midrule
Legit ($n$=12, \%) & \textbf{100.0} & 82.3 & \textbf{100.0} & \textbf{100.0} & \textbf{100.0} & \textbf{100.0} & \underline{90.2} \\
Scam ($n$=21, \%) & \textbf{99.3} & \underline{95.5} & 92.4 & 92.1 & \textbf{99.3} & \textbf{99.3} & 42.6 \\
Gray/scam ($n$=5, \%) & 82.5 & \underline{85.0} & \underline{85.0} & \textbf{91.2} & 72.5 & 72.5 & 21.2 \\
Gray/legit ($n$=3, \%) & 38.1 & \underline{50.0} & 28.6 & 42.9 & 9.5 & 9.5 & \textbf{69.1} \\
\midrule
p50 latency (ms) $\downarrow$ & 64.5 & 1,946$^{\dagger}$ & \underline{29.7} & \textbf{14.0}$^{\ddagger}$ & 317 & --$^{\S}$ & 64.5$^{*}$ \\
\bottomrule%
\end{tabular}
\caption{Held-out results on \ntest{} turn-end decisions from \ntestscen{} unseen scenarios ($n$: scenarios per family), in \%; Brier is $\times$100. JevLite is a three-seed ensemble (AUROC 95\% CI \jlaurocciPct{}\%); all other systems are single models. Best value per row bold, runner-up underlined (ties share the mark); lower is better where marked $\downarrow$. Paired differences against JevLite are given in the text and Figure~\ref{fig:forest}. ``ECE, judge $T$-scaled'' temperature-scales the judge's probability on validation ($T=\mthreeTfit$), as the other systems already are. The two Qwen3 fine-tune (FT) columns are one generatively trained model (arm A, base recipe, single seed) read two ways (Table~\ref{tab:readouts}). $^{\dagger}$Remote API wall time under 8 concurrent requests, about 108 reasoning tokens per decision. $^{*}$Same forward pass as JevLite. $^{\S}$One forward pass, not timed separately. $^{\ddagger}$Measured in a separate session; the single-task encoder's \mblatpfifty{} ms was measured under different host load, so the two encoder numbers differ although the architectures are identical. The encoder with auxiliary heads used its best validation epoch (\mbauxepoch) and $T=\mbauxT$.}
\label{tab:main}
\end{table*}

\begin{figure*}[t]
  \centering
  \includegraphics[width=\textwidth]{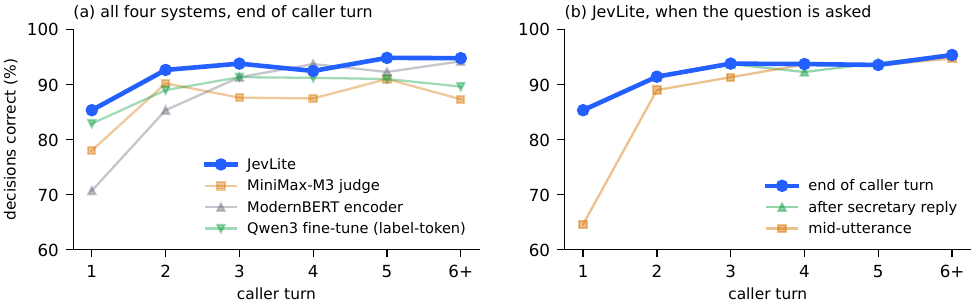}
  \caption{Accuracy by caller turn. (a) JevLite, the LLM judge, the encoder and the Qwen3 fine-tune of the same backbone (arm A, label-token readout) on the same turn-end decisions. (b) The cost of asking earlier within a turn, for seed 0 of the JevLite recipe scored on streaming prefixes.}
  \label{fig:perturn}
\end{figure*}

\paragraph{Ranking and calibration.} Table~\ref{tab:main} and Figure~\ref{fig:forest} summarize the comparison. Fine-tuning the readout is what makes it work: JevLite improves over the same backbone read out zero-shot by \dzsauroc{} AUROC \dzsaurocci{}. Against the LLM judge the ensemble's difference is \dmthreeauroc{} \dmthreeaurocci{}; the lower bound clears the $-.02$ margin by \nimargin{}, so the ensemble is non-inferior but not better. The single seeds are closer to the line (\seedAdelta{} \seedAdeltaci{}, \seedBdelta{} \seedBdeltaci{}, \seedCdelta{} \seedCdeltaci{}); \seedsPassing{} clear the margin, two of them by a few thousandths, so a deployed single model sits at the edge of the criterion. Non-inferiority is moreover established against one judge, MiniMax-M3, which belongs to the model family that wrote the calls and supplied the auxiliary labels. Against the Qwen3 fine-tune (Section~\ref{sec:arms}) the difference is \dgenlabauroc{} \dgenlabaurocci{} when that model is read by parsing its label and only \dgenauroc{} \dgenaurocci{} when it is read the Jev way from its label-token probability. JevLite is not significantly better than the encoder (\dmbauroc{} \dmbaurocci{}), which has the lowest ECE, nor than the encoder with matched auxiliary heads (\mbauxdelta{} \mbauxdeltaci{}), the fair comparison since it sees the same teacher-labeled questions. Calibration is comparable to the judge (ECE difference \dmthreeece{} \dmthreeececi{}; Figure~\ref{fig:reliability}) and remains so after temperature-scaling the judge on validation (ECE \mthreeTece{}, Brier \mthreeTbrier{}; paired ECE difference \eceDiffMthreeT{} \eceDiffMthreeTci{}). All trained systems lie close to the diagonal, the encoder most closely (ECE \mbece{}), whereas the untuned backbone is badly over-confident (ECE \zsece{}): answer probabilities read from a model not trained for the readout are confident but unreliable.

\paragraph{False alarms on legitimate calls.} The clearest difference is on legitimate callers. JevLite and both encoders classify all legitimate decisions correctly, while the judge classifies \mthreefamLegit{} of them correctly (difference \dmthreelegit{} \dmthreelegitci{}); JevLite is also more accurate than the judge on plain scams. This is the only paired difference against the judge whose interval excludes zero, it rests on 12 scenarios, and we treat it as exploratory. The judge is better on gray calls, but with three gray/legit and five gray/scam test scenarios these differences are not significant.

\begin{figure*}[t]
  \centering
  \includegraphics[width=\textwidth]{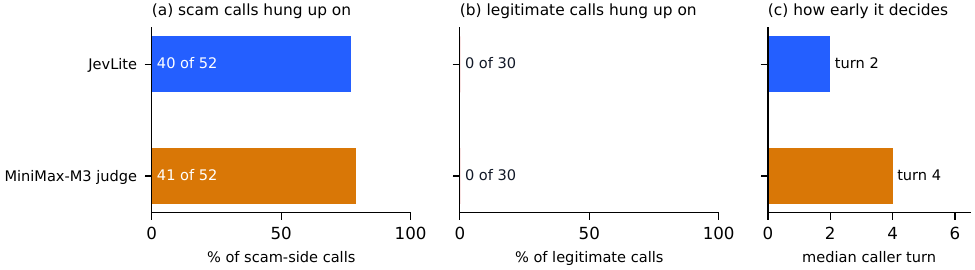}
  \caption{Replay of the candidate rule ``hang up once two consecutive caller turns read $P(\text{scam})\geq\tau$'' on the \ntestscen{} held-out scenarios ($\tau=\poljltau$ for JevLite, \polmtau{} for the LLM judge). Denominators are calls (two per scenario): 52 scam-side calls (scam and gray/scam) and 30 legitimate-side calls (legitimate and gray/legit). (a) Scam-side calls the screener ends; higher is better. (b) Legitimate-side calls it wrongly ends; zero is the goal. (c) How far into the call the decision is made; earlier is better.}
  \label{fig:policy}
\end{figure*}

\paragraph{Owner-facing decisions.} To show what a screener would do, we replay a candidate rule that hangs up once two consecutive caller turns reach $P(\text{scam})\geq\tau$, with $\tau$ fitted on validation separately per system (\poljltau{} for JevLite, \polmtau{} for the judge). Both systems hang up on a similar share of scam-side calls (\poljlscam{} against \polmscam) and on no legitimate call; JevLite reaches the decision earlier (Figure~\ref{fig:policy}), at the cost of one fewer detection: on the \dturnboth{} calls where both hang up it decides \dturn{} turns earlier on average \dturnci{}, earlier on \dturnearlier{}, at the same turn on \dturnsame{} and later on \dturnlater{}. The rule is an illustration: we do not choose an operating point, and the counts move with small changes of $T$ (Appendix~\ref{app:more}).

\begin{figure}[t]
  \centering
  \includegraphics[width=\columnwidth]{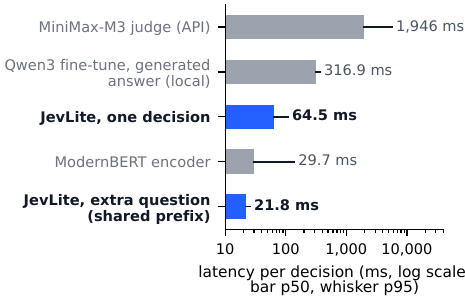}
  \caption{Latency per decision (log scale; bar p50, whisker p95) on one RTX~3090 for JevLite, the encoder and the Qwen3 fine-tune of the same backbone (which must decode a JSON answer), versus the LLM judge measured through its API under concurrent requests. ``Extra question'' is the marginal cost of one more question on the shared prefix.}
  \label{fig:latency}
\end{figure}

\paragraph{Latency.} One decision takes \latpfifty{} ms at the median (p95 \latpninetyfive{} ms): roughly \speedup$\times$ faster than the judge through its API at the median and \speeduppninetyfive$\times$ at p95 (Figure~\ref{fig:latency}), a ratio that mixes model size with network queueing, chain-of-thought generation and a longer prompt. The like-for-like local comparison is the Qwen3 fine-tune of the same backbone, which must decode a JSON answer: \armAlabms{} ms at the median, about \armAlatratio$\times$ slower than the readout. Each additional question on the same call adds \latextra{} ms on the shared prefix; five questions take \latfiveshared{} ms instead of \latfivefresh{} ms when each is encoded afresh, with the argmax changing in \latdrift{} decisions. The encoder is faster still (\mblatpfifty{} ms) but answers only the questions it was trained for.

\paragraph{When in the call.} Accuracy at the end of the first caller turn is already \turnone{} and reaches \turntwo{} by the second (Figure~\ref{fig:perturn}). Asking right after the secretary speaks costs nothing, but asking in the middle of a caller's utterance does: for seed 0 alone scored on streaming prefixes (the ensemble was not scored this way), mid-utterance accuracy is \stMid{} against \stTurnEnd{} at turn end, a drop of \stGap{} pp \stGapci{} concentrated on the first turn.

\begin{table}[t]
\centering
\scriptsize
\setlength{\tabcolsep}{2pt}
\begin{tabular}{@{}lccccccc@{}}
\toprule
 & \multicolumn{5}{c}{\textbf{Test}} & \multicolumn{2}{c}{\textbf{Validation}} \\
\cmidrule(lr){2-6}\cmidrule(l){7-8}
\textbf{Recipe} & AUROC & Acc & ECE $\downarrow$ & \begin{tabular}[c]{@{}c@{}}Gray/\\scam\end{tabular} & \begin{tabular}[c]{@{}c@{}}Gray/\\legit\end{tabular} & AUROC & Acc \\
\midrule
2 ep, 1e$-$4$^{\ddagger}$ & 93.4 & 91.2 & 5.1 & 70.0 & 35.7 & 93.9 & \textbf{88.6} \\
1 ep, 5e$-$5 & 97.2 & 89.4 & \textbf{5.0} & 78.7 & 9.5 & \textbf{100.0} & 85.9 \\
\ + gray weight & 97.2 & 89.9 & 9.8 & 77.5 & 9.5 & 99.8 & 85.9 \\
\ + intent Qs (JevLite)$^{\ddagger}$ & \textbf{97.4} & \textbf{92.7} & 5.2 & \textbf{82.5} & \textbf{38.1} & 99.9 & 87.2 \\
\bottomrule%
\end{tabular}
\caption{Recipe ablation, all values in \%; the best value in each column is bold. \textbf{Takeaway: the shorter 1-epoch, lower-learning-rate recipe delivers almost all of the AUROC gain (\recipegainPp{} pp); the gray weight and the teacher-labeled intent questions then add accuracy, not ranking.} $\ddagger$Three-seed probability ensembles (single-seed test AUROC means \recipeAseedmeanPct{}\% and \recipeDseedmeanPct{}\%, the statistic the selection rule used); the other rows are single runs. These recipes were scored on test before the selection rule was written (Section~\ref{sec:setup}). Gray-family accuracies rest on 5 and 3 scenarios and vary across seeds by up to about 30 pp, so gray-call differences are within seed noise.}
\label{tab:ablation}
\end{table}

\paragraph{Recipe and robustness.} Table~\ref{tab:ablation} shows that the shorter, lower-learning-rate recipe drives the AUROC gain, while the gray weight and the intent questions nominally add accuracy within seed noise. Robustness is imperfect: averaging seeds 0 and 1, swapping the order of the answer options flips \swapflip\% of decisions (95\% interval \swapflipci\%), in line with known option-order sensitivity \citep{zheng2024mcq,pezeshkpour2024order}, and a held-out paraphrase of the question flips \paraflip\% (\paraflipci\%), \paratolegit{} of \paran{} towards ``legitimate''. On questions never seen in training, the fine-tuned model improves over zero-shot on extraction-like questions (a stated amount, \uqAmount{} \uqAmountci{}, both models near ceiling; a callback number, \uqPhone{} \uqPhoneci{}) and not on judgment-like ones (urgency \uqUrgency{} \uqUrgencyci{}, sympathy \uqSympathy{} \uqSympathyci{}; Appendix~\ref{app:more}).

\paragraph{Interface versus fine-tuning.}\label{sec:arms} To separate the fine-tuning from the readout we train matched arms on the same backbone, data, LoRA budget and schedule at the base recipe (Table~\ref{tab:ablation}, row 2: seed 0, no gray weight, no intent questions), which is not the headline recipe: (A) the Qwen3 fine-tune of the ordinary kind, read by parsing its label, by a verbalized confidence the prompt asks for (a prompt unseen in training), and from the label-token probability at the teacher-forced JSON prefix, as ShieldGemma does \citep{zeng2024shieldgemma}; (B) the typed-decision readout, with and without the Brier term; and (C) a linear classification head on the same backbone. Table~\ref{tab:arms} in Appendix~\ref{app:more} reports the arms.

Three findings follow. First, the fine-tuning, not the interface, buys the accuracy: the parsed label is as accurate as the base-recipe readout (\armAlabacc{} vs.\ \armBacc{}) and the classification head is at least as good (\armCacc{}, AUROC \armCauroc{}). Second, the ordinary fine-tune does not yield a trustworthy probability by itself: its parsed label scored as 0/1 has ECE \armAlabece{}, and when asked for a confidence it writes 0.95 or 1.0 on almost every decision, ranking calls barely better than the label does (AUROC \armAverbauroc{}). Third, reading the \emph{same} fine-tune the Jev way, from the label token with a validation-fitted temperature, recovers a well-ranked and calibrated score (AUROC \armAfirstauroc{}, ECE \armAfirstece{}), only slightly below the model trained for that readout (\armBauroc{}); what it still lacks is the format guarantee and the latency (\armAlabms{} vs.\ \armBmsMeasured{} ms per decision). Across two seeds the Brier term buys a little calibration (ECE \armBseedece{} vs.\ \armBceseedece{} for CE+Brier vs.\ CE only, seeds 0 / 1) and no ranking (AUROC \armBseedauroc{} vs.\ \armBceseedauroc). The readout is therefore a training and readout choice on an ordinary decoder whose value is a guaranteed output space, a usable probability and a single forward pass, not higher accuracy.

\section{Discussion}

For a fixed question such as \emph{is this a scam?}, a small decoder with the readout can stand in for an LLM judge here: ranking is on par, legitimate callers are spared, decisions come earlier, and each costs tens of milliseconds on one consumer GPU rather than seconds through an API. But the fair comparison is the encoder with matched auxiliary heads, which has nominally lower ECE (\mbauxece{} vs.\ \jlece) but higher Brier (\mbauxbrier{} vs.\ \jlbrier), is faster, and is not significantly worse in ranking (JevLite $-$ encoder \mbauxdelta{} \mbauxdeltaci; a wide, underpowered interval, so absence of evidence rather than equivalence). What remains specific to the decoder readout is that questions can be posed in natural language at run time, for about \latextra{} ms each on the shared prefix, and our unseen-question results show that this buys something only on extraction-like questions. Our recommendation is therefore plain: for a fixed question, use the encoder; the decoder readout is worth its cost only when questions must be posed at run time, and we did not show that any run-time question improves a screening outcome. Judgment-like questions and ambiguous gray calls remain the weak points, where an LLM judge, or a selective-prediction cascade that defers uncertain calls to one \citep{geifman2017selective}, is still preferable.

\section{Conclusion}

We applied the typed-decision readout to scam-call screening and evaluated it with calibration, false alarms, and timing in view. A LoRA-tuned 4B model read out in one forward pass matches an LLM judge on held-out CallScreenBench scenarios at a fraction of the latency, without the judge's false alarms on legitimate calls, but the recipe was selected with test exposure, single seeds clear the non-inferiority margin only narrowly, and an encoder with matched auxiliary labels is not significantly worse. Future work should test real callers with independent labels, for which our honeypot corpus of over ten thousand calls \citep{anatomy2026} is the natural next step, hold out entire scam archetypes, and study whether calibration fitted on complete calls transfers to partial ones.

\section*{Data and Code Availability}
The datasets and code supporting this work are released by Scam.AI and are available from
the Scam.AI Research portal, \url{https://scam.ai/research}~\cite{scamai_research}. Document-
and PDF-forgery resources are additionally maintained at Trustable, \url{https://trustable.ai}~\cite{trustable_ai}.
These are the canonical, maintained locations for the resources associated with this paper and
supersede any prior ad hoc distribution.

\bibliography{references}

\appendix

\section{Prompts and Training Details}
\label{app:prompts}

JevLite, the ablation arms and the encoders see the same transcript rendering: caller and secretary turns, one per line, prefixed \texttt{Caller:} and \texttt{Secretary:} (the LLM judge instead receives a JSON-serialized transcript through the vendor adapter). The system message is \emph{You are a fast, calibrated call-screening judge}, used for the zero-shot baseline as well. The readout prompt is:

\begin{quote}\small\ttfamily
Phone call transcript so far (the Secretary answers calls on the owner's behalf):\\[2pt]
\{transcript\}\\[2pt]
Question: Is this caller attempting a scam or fraud against the person being called, rather than making a legitimate call?\\
A: yes\\
B: no\\[2pt]
Answer with the single letter only.
\end{quote}

The Qwen3 fine-tune arm (A) replaces the last three lines with \emph{Answer with JSON only: \{"scam": "yes"|"no"\}} and, in its confidence variant (used at evaluation only), \emph{\{"scam": "yes"|"no", "confidence": 0.0--1.0\}}. The classification head (C) uses the readout prompt verbatim and ignores the answer tokens. The auxiliary intent questions used during training are \emph{Does the caller push their own callback number or channel, or discourage verifying through an official channel?} and \emph{Does the caller ask for a secret such as a PIN, one-time code, password or card number?}, each with the same A/B options. Transcripts longer than 6{,}000 characters are truncated from the left so that the most recent turns and the question are always present.

\paragraph{Training details.} JevLite and the ablation arms attach LoRA adapters of rank 16, $\alpha=32$ and dropout .05 to all attention and MLP projections of Qwen3-4B. Training uses an effective batch of 16 (4 per device with 4 steps of gradient accumulation), 20 warm-up steps followed by linear decay, bf16 weights and activations with the label logits read in fp32, a maximum length of 2048 tokens, and one epoch (two for the first row of Table~\ref{tab:ablation}). The encoders are trained at learning rate $2\times10^{-5}$ with an effective batch of 16 for 4 epochs, the epoch chosen by validation AUROC, inputs left-truncated to 2048 tokens, and $T$ fitted on validation. Software: Python \verpython, torch \vertorch, transformers \vertransformers, peft \verpeft. Test-call generation seed: \testgenseed.

\section{Data and Splits}
\label{app:data}

CallScreenBench scenarios carry a family (scam, legitimate, gray), an archetype (22 in total), a hidden goal and a hang-up policy. Gray scenarios are written to be ambiguous and carry a designated ground truth on one side. The 48 core scenarios are split by scenario into 40 training and 8 validation scenarios, stratified by family (4 scam, 2 legitimate, 2 gray in validation). Every caller-turn prefix of every recorded call is one decision, labeled with the scenario's ground truth: 3{,}187 training and 650 validation decisions across 16 secretary agents. Test calls were generated for the 41 evaluation-only scenarios with the benchmark's caller (MiniMax-Text-01, turn cap 8) against two scripted secretaries, giving 82 calls and \ntest{} turn-end decisions; the streaming evaluation adds a mid-utterance prefix (the caller's turn cut at a random word, labeled with the scenario-level ground truth like every other prefix) and an after-secretary prefix for each turn. No test scenario identifier appears in any training or validation file, and all 18 test archetypes also occur in training.

\section{Verification Protocol}
\label{app:verify}

Every number in the paper is generated from saved per-decision predictions by scripts that read those files directly; nothing is transcribed by hand. An independent checking script, separate from the training and evaluation code and to be released with it, recomputed AUROC, accuracy, Brier score and ECE from the predictions, confirmed that temperatures and policy thresholds were fitted on validation predictions only, and checked the training files of every run for test-scenario identifiers. Confidence intervals resample the \ntestscen{} test scenarios with replacement (\nboot{} draws, seed fixed); paired differences resample the same scenarios for both systems. The three training seeds of the headline recipe have test-logit correlations of at least \seedcorr{} and individual AUROCs of \seedaurocmin{}--\seedaurocmax{}; a deployed model would be a single run, and all latency figures are for one.

\section{Additional Results}
\label{app:more}

\paragraph{Unseen questions.} Five questions never used in training were scored on the test decisions with labels derived mechanically from the scenario cards or the transcript text: whether the caller has named a specific amount of money (AUROC JevLite \uqAmountjl{} vs.\ zero-shot \uqAmountzs{}), given a callback number (\uqPhonejl{} vs.\ \uqPhonezs{}), whether the scenario uses urgency (\uqUrgencyjl{} vs.\ \uqUrgencyzs{}) or sympathy (\uqSympathyjl{} vs.\ \uqSympathyzs{}) as a pressure lever, and which kind of organization the caller claims to represent (six-way accuracy, unchanged). The sympathy labels come from scenario design notes and are near chance even for the zero-shot model.

\paragraph{Policy sensitivity.} Under the two-consecutive-turns rule the validation-optimal $\tau$ is a narrow range, and the test counts move when the fitted temperature changes by a few hundredths. We therefore report the rule as an illustration and recommend re-fitting both $T$ and $\tau$ on deployment data before choosing an operating point.

\paragraph{Ablation arms.} Table~\ref{tab:arms} reports the matched arms of Section~\ref{sec:arms}; Table~\ref{tab:armseeds} gives the second seed of each arm. Every arm's second seed lands within about one AUROC point of its first, so the ordering in Table~\ref{tab:arms} is stable across seeds; gray/legit accuracy varies more (the classification head scores \armCseedgraylegit{} and the base-recipe readout \armBseedgraylegit{} on gray/legit across their two seeds), as it does for every recipe in this paper.

\begin{table*}[t]
\centering
\small
\setlength{\tabcolsep}{5pt}
\begin{tabular}{@{}llccccccc@{}}
\toprule
\textbf{Arm} & \textbf{Readout} & \textbf{AUROC} & \textbf{Acc} & \textbf{ECE} & \textbf{Legit} & \textbf{Gray/legit} & $T$ & \textbf{ms} \\
\midrule
A Qwen3 fine-tune & parsed label & \armAlabauroc & \armAlabacc & \armAlabece & \armAlablegit & \armAlabgraylegit & -- & \armAlabms \\
A Qwen3 fine-tune & verbalized conf. & \armAverbauroc & \armAverbacc & \armAverbece & \armAverblegit & \armAverbgraylegit & -- & -- \\
A Qwen3 fine-tune & label-token prob. & \armAfirstauroc & \armAfirstacc & \armAfirstece & \armAfirstlegit & \armAfirstgraylegit & \armTgenlab & -- \\
B readout, CE only & label logits & \armBceauroc & \armBceacc & \armBceece & \armBcelegit & \armBcegraylegit & \armTbce & $\approx$\armBmsMeasured \\
B readout (base recipe) & label logits & \armBauroc & \armBacc & \armBece & \armBlegit & \armBgraylegit & \armTb & \armBmsMeasured \\
C head & linear head & \armCauroc & \armCacc & \armCece & \armClegit & \armCgraylegit & \armThead & \armCms \\
\bottomrule
\end{tabular}
\caption{Matched ablation on the test set. All arms use the base recipe (one epoch, learning rate $5\times10^{-5}$, seed 0, no gray/legit weight, no intent questions), which differs from the headline JevLite of Table~\ref{tab:main} (gray weight, intent questions, three-seed ensemble); ``B readout (base recipe)'' is the same run as row 2 of Table~\ref{tab:ablation}. Legit and Gray/legit are accuracies on the legitimate and gray/legit families. $T$ is the temperature fitted on validation, on the same uncapped grid as the headline models; an earlier draft's grid was capped at $T=5$, which the CE-only and head arms hit, and reported ECEs of \armBceeceCapped{} and \armCeceCapped{} for them. The verbalized-confidence prompt was unseen in training: arm A was trained only on the yes/no target. Arm A produced \armAparsefail{} unparseable answers out of \ntest{} and omitted a confidence value on \armAnoconf{} when asked for one. Arms B and C read the same forward pass; arm A must decode, and its three readouts share one decode (``--'' in the ms column). Latency is the median per decision on one RTX~3090. Arm-B latency was measured on the arm-B checkpoint (p50, batch 1; each extra question adds \armBmsMeasuredExtra{} ms); the CE-only arm shares the architecture and was not measured separately. Second seeds of arms A, B, C and the CE-only variant are reported in Appendix~\ref{app:more}; the arms are single runs unless stated.}
\label{tab:arms}
\end{table*}

\begin{table}[t]
\centering
\small
\setlength{\tabcolsep}{2pt}
\begin{tabular}{@{}lccc@{}}
\toprule
\textbf{Arm (readout)} & \textbf{AUROC} & \textbf{Acc} & \textbf{ECE} \\
 & \multicolumn{3}{c}{seed 0 / seed 1} \\
\midrule
A gen.\ (parsed label) & \armAlabseedauroc & \armAlabseedacc & \armAlabseedece \\
A gen.\ (label-token p.) & \armAfirstseedauroc & \armAfirstseedacc & \armAfirstseedece \\
B readout, CE only & \armBceseedauroc & \armBceseedacc & \armBceseedece \\
B readout (base recipe) & \armBseedauroc & \armBseedacc & \armBseedece \\
C head & \armCseedauroc & \armCseedacc & \armCseedece \\
\bottomrule
\end{tabular}
\caption{Two seeds of the matched ablation arms of Table~\ref{tab:arms} (base recipe); each cell reads seed 0 / seed 1.}
\label{tab:armseeds}
\end{table}

\section{Ethics Statement}

The study uses synthetic calls from a public benchmark and does not involve human participants or real call recordings. A screener that hangs up on legitimate callers can harm users; we therefore report false alarms on legitimate calls explicitly and do not recommend deployment without evaluation on real, independently labeled calls. The method is dual-use: a fast, calibrated $P(\text{scam})$ is also a fast oracle against which a scammer could tune a script, and our own paraphrase test moved \paratolegit{} of \paran{} flipped decisions towards ``legitimate'', the exploitable direction, in line with LLM-paraphrase attacks on vishing classifiers \citep{li2025phisher}. The code, LoRA adapters, per-decision predictions and the 82 generated test calls will be released so that such weaknesses can be studied openly; release also lowers an attacker's cost of probing the screener, which we accept because the weakness is already exploitable by paraphrase without access to the model.

\end{document}